\documentclass{article} 

\usepackage{amssymb}
\usepackage{graphicx}
\usepackage[figuresright]{rotating}
\usepackage{tabularx}
\usepackage{color}
\usepackage{soul}
\usepackage{epstopdf}
\usepackage[utf8]{inputenc}
\usepackage[]{siunitx}
\usepackage{hyperref}
\usepackage{xstring}
\usepackage{cleveref}
\usepackage[authoryear, sort]{natbib}
\usepackage{caption}

\DeclareCaptionLabelFormat{andtable}{#1~#2  \&  \tablename~\thetable}

\renewcommand{\hl}[1]{\ }

\begin{document}

\title{Cross-lingual alignments of ELMo contextual embeddings}

\author{Matej Ul\v{c}ar, Marko Robnik-\v{S}ikonja \\
University of Ljubljana, Faculty of Computer and Information Science \\
Ve\v{c}na pot 113, Ljubljana, Slovenia \\
\{matej.ulcar, marko.robnik\}@fri.uni-lj.si
}

\maketitle
\begin{abstract}
 Building machine learning prediction models for a specific NLP task requires sufficient training data, which can be difficult to obtain for less-resourced languages. Cross-lingual embeddings map word embeddings from a less-resourced language to a resource-rich language so that a prediction model trained on data from the resource-rich language can also be used in the less-resourced language. 
To produce cross-lingual mappings of recent contextual embeddings, anchor points between the embedding spaces have to be words in the same context. We address this issue with a novel method for creating cross-lingual contextual alignment datasets. Based on that, we propose several cross-lingual mapping methods for ELMo embeddings. The proposed linear mapping methods use existing Vecmap and MUSE alignments on contextual ELMo embeddings. Novel nonlinear ELMoGAN mapping methods are based on GANs and do not assume isomorphic embedding spaces. We evaluate the proposed mapping methods on nine languages, using four downstream tasks: named entity recognition (NER), dependency parsing (DP), terminology alignment, and sentiment analysis. The ELMoGAN methods perform very well on the NER and terminology alignment tasks, with a lower cross-lingual loss for NER compared to the direct training on some languages. In DP and sentiment analysis, linear contextual alignment variants are more successful. 
\end{abstract}

\section{Introduction}
\label{sec:intro}
Word embeddings are representations of words in a numerical form, as vectors of typically several hundred dimensions. The vectors are used as inputs to machine learning models; these are generally deep neural networks for complex language processing tasks. The embedding vectors are obtained from specialized neural network-based embedding algorithms. The quality of embeddings depends on the amount of semantic information expressed in the embedded space through distances and directions. For that reason, static pre-trained word embeddings, such as word2vec \cite{mikolov2013exploiting} or fastText \cite{Bojanowski2017}, have in large part been recently replaced by contextual embeddings, such as ELMo \cite{Peters2018} and BERT \cite{Devlin2019}. 

Contextual embeddings generate a different word vector for the same word for every context it appears in. BERT models and their derivatives are mostly used as a closed system, where the entire model is fine-tuned on a downstream task. On the other hand, ELMo models generate different word vectors for each word occurrence, and these vectors are used in training NLP models. A neural network producing ELMo embeddings contains three layers of neurons. Embeddings are typically a concatenation of network weights in all three layers. BERT models consist of 12 or 24 layers, and vector extraction typically uses a combination of the last four layers. Due to the omission of most network layers, explicit BERT vectors may lack a lot of information, For that reason, the explicit BERT vectors are rarely used and are often less successful than ELMo vectors, see, e.g., \citep{skvorc2020MICE}. A smaller size of ELMo models compared to BERT, may also offer better explainability of the end-task models.

Modern word embedding spaces exhibit similar structures across languages, even when considering distant language pairs like English and Vietnamese \citep{mikolov2013exploiting}. This means that embeddings independently produced from monolingual text resources can be aligned, resulting in a common cross-lingual representation, called cross-lingual embeddings, which allows for fast and effective integration of information in different languages. 
For less-resourced languages, training NLP models can be difficult because of a lack of data for a specific task. The aim of cross-lingual alignment is to use an already existing model trained on a resource-rich language and map the word embeddings from a less-resourced language vector space to the resource-rich language vector space. In that way, the input in less-resourced language is mapped to resource-rich language and can be classified with existing models in that language. This is possible as the words with the same meaning in both languages have very similar vectors after the cross-lingual alignment.  

Cross-lingual approaches can be sorted into several groups. The first group of methods uses monolingual embeddings with (an optional) help from bilingual dictionaries to align the embeddings. These methods are typically used for static embeddings, such as word2vec and fastText.  The second group of approaches uses bilingually aligned (comparable or even parallel) corpora for joint construction of embeddings in all involved languages. An example of this approach is a joint space of 93 languages produced by LASER library \citep{artetxe2019massively}. The third type of approach is based on large pretrained multilingual masked language models such as BERT \citep{Devlin2019}, simultaneously trained on many languages and not needing explicit mappings. In this work, we present an extension of the first group of approaches to contextual embeddings. We focus on improvements of cross-lingual mappings for ELMo contextual embeddings (as BERT does not need them). Currently, the most successful alignment methods assume that the embedding spaces in different languages are isomorphic \citep{artetxe2018robust, Conneau2018}, which is generally not the case. Researchers have observed that the monolingual embedding spaces of two different languages are not completely isomorphic, which is especially true for distant languages \citep{ormazabal-etal-2019-analyzing,sogaard2018limitations}. As a result, many of these methods are unstable or unsuccessful when confronted with distant language pairs. 

We propose novel methods for linear and non-linear alignment of contextual embeddings, such as ELMo. For that purpose, we first construct novel contextual mapping datasets based on parallel corpora and dictionaries. In the novel ELMoGAN approach, we use generative adversarial networks (GANs) \citep{goodfellow2014generative}, that produce nonlinear mappings between the embedding spaces. 
The main contributions of this work are as follows.
\begin{enumerate}
    \item A novel approach to create datasets needed in the cross-lingual alignment of contextual embeddings.
    \item Novel linear and non-linear cross-lingual alignment methods for ELMo embeddings.
    \item Evaluation of produced cross-lingual embeddings using nine languages and four downstream tasks: named entity recognition (NER), dependency parsing (DP), terminology alignment, and sentiment analysis. 
\end{enumerate}
The results show a successful cross-lingual transfer of tested approaches. The best alignment method depends on the task. We publish the code to create the needed datasets and the code to produce non-linear cross-lingual alignment under a permissive license.\footnote{\url{https://github.com/MatejUlcar/elmogan}}

The paper is split into five further sections. In Section~\ref{sec:relatedwork}, we present the background on cross-lingual alignment and ELMo and cover related work on cross-lingual embeddings. The construction of special datasets used for training the alignments of contextual embeddings is presented in \Cref{sec:alignmentDatasets}. In Section~\ref{sec:elmogan}, we describe the proposed non-linear ELMoGAN alignment method and linear methods combining the new dataset and existing mapping methods. In Section~\ref{sec:evaluation}, we evaluate the proposed alignment methods on four downstream tasks. We summarize our work in Section~\ref{sec:conclusions} and discuss opportunities for further work.

\section{Background and related work}
\label{sec:relatedwork}

\label{sec:XLbackground}

Word embeddings represent each word in a language as a vector in a high dimensional vector space so that the relations between words in a language are reflected in their corresponding embeddings. Cross-lingual embeddings attempt to map words represented as vectors from one vector space to the other so that vectors representing words with the same meaning in both languages are as close as possible. \citet{Sogaard2019} present a detailed overview and classification of cross-lingual methods.

In Section  \ref{sec:monoAlignments}, we describe how two monolingual embedding spaces can be aligned with the optional help from a bilingual dictionary. This work's main focus is extending existing approaches that work with non-contextual embeddings to contextual ELMo embeddings. For this reason, we present the background on ELMo contextual embeddings in \Cref{sec:elmo}. The related work on non-contextual mappings is given in \Cref{sec:relatedStatic}, and on contextual mapping in \Cref{sec:relatedContextual}.

\subsection{Alignment of monolingual embeddings}
\label{sec:monoAlignments}
Cross-lingual alignment methods take precomputed word embeddings for each language and align them with the optional use of bilingual dictionaries. Two types of monolingual embedding alignment methods exist. The methods of the first type map vectors representing words in one of the languages into the vector space of the other language. 
The methods of the second type map embeddings from both languages into a common vector space. The goal of both types of alignments is the same: the embeddings for words with the same meaning must be as close as possible in the final vector space. A comprehensive summary of existing approaches can be found in works by \citet{artetxe2018generalizing}.
The open source implementation of the method described by \citet{artetxe2018robust,artetxe2018generalizing}, named \emph{Vecmap}\footnote{\url{https://github.com/artetxem/vecmap}}, is able to align monolingual embeddings using supervised, semi-supervised or unsupervised approach.
    
The supervised approach requires a large bilingual dictionary, which is used to match embeddings of the words with the same meaning. The embeddings are aligned using the Moore-Penrose pseudo-inverse, which minimizes the sum of squared Euclidean distances.
The algorithm always converges but can be caught in a local maximum when the initial solution is poor. To overcome this, several methods (stochastic dictionary introduction, frequency-based vocabulary cutoff, etc.) are used that help the algorithm to climb out of local maximums. A more detailed description of the algorithm is given in \citep{artetxe2018robust}.

The semi-supervised approach uses a small initial seeding dictionary, while the unsupervised approach is run without any bilingual information. The latter uses similarity matrices of both embeddings to build an initial dictionary. This initial dictionary is usually of poor but sufficient quality for later processing.   
After the initial dictionary (either by seeding dictionary or using similarity matrices) is built, the iterative algorithm is applied. The algorithm first computes an optimal mapping using the pseudo-inverse approach for the given initial dictionary. Then optimal dictionary for the given embeddings is computed, and the procedure is repeated with the new dictionary.


Besides \emph{Vecmap}, another well-known library for cross-lingual embeddings alignment is called MUSE\footnote{\url{https://github.com/facebookresearch/MUSE}}. This library can find a cross-lingual map with the use of a bilingual dictionary (supervised) or without one (unsupervised approach). The unsupervised approach works by using adversarial training to find the starting linear mapping. A synthetic dictionary is extracted from this mapping, which is used to fine-tune the starting mapping using the Procrustes approach, described in detail by \citet{Conneau2018}.

\subsection{ELMo contextual embeddings} 
\label{sec:elmo}
ELMo (Embeddings from Language Models) embedding \citep{Peters2018} is an example of a state-of-the-art pre-trained contextual embedding model. It is a neural network model, composed of three layers.
The first layer is a CNN layer, which operates on a character level. It is context-independent, so each word always gets the same embedding from this layer, regardless of its context. It is followed by two biLM (bidirectional language model) layers. A biLM layer consists of two concatenated LSTMs \citep{hochreiter1997-lstm}. In the first LSTM, the network predicts the following word, based on the given past words, where the embeddings from the CNN layer represent each word. In the second LSTM, the network predicts the preceding word based on the given following words. This layer is equivalent to the first LSTM, just reading the text in reverse.

The actual embeddings are constructed from the internal states of the bidirectional LSTM neural network. Two higher-level LSTM layers capture context-dependent aspects, while the first CNN layer captures aspects of syntax \citep{Peters2018}. 
To train the ELMo network, one puts one sentence at a time on the input. The representation of each word depends on the whole sentence, i.e. it reflects the contextual features of the input text and thereby polysemy of words. For an explicit word representation, one can use only the top layer. Still, more frequently, one combines all layers into a vector.
The representation of a word or a token $t_k$ at position $k$ is composed of
\begin{equation}
R_k  =  \{ x_k^{LM}, \overrightarrow{h}_{k,j}^{LM}, \overleftarrow{h}_{k,j}^{LM} ~|~ j=1,\dots,L\}  
\end{equation}
where $L$ is the number of layers (ELMo uses $L=2$), index $j$ refers to the level of bidirectional LSTM network, $x$ is the initial token representation (either word or character embedding), and $h^{LM}$ denotes hidden layers of a forward or a backward language model. 
In NLP tasks, a weighted average of layers is usually used, where the weights are learned during the training of the model for the specific task. Alternatively, the entire ELMo model can be fine-tuned on a specific end-task.

At the time of its introduction, ELMo has been shown to outperform previous pre-trained word embeddings like word2vec and GloVe on many NLP tasks, e.g., question answering, named entity extraction, sentiment analysis, textual entailment, semantic role labelling, and coreference resolution \citep{Peters2018}. Later, BERT models turned out to be even more successful on these tasks. However, concerning the quality of extracted vectors, ELMo can be advantageous as its information is condensed in only three layers. In comparison, the information in multilingual BERT is scattered over 14 layers and a reasonably sized embedding drops most of them.

\subsection{Related work on non-contextual mappings}
\label{sec:relatedStatic}
Cross-lingual alignment methods align precomputed monolingual word embeddings from two or more languages. The word vectors from all the languages are mapped into a common vector space. This can be the same vector space as one of the original monolingual embeddings or a completely independent vector space. These methods aim to represent the words with the same meaning in different languages with as similar vectors as possible. Concerning the data used, the alignment methods can be split into supervised and unsupervised methods. Supervised methods determine the alignment of the embeddings with the use of bilingual dictionaries. Unsupervised methods do not use any bilingual data. \citet{conneau2018word} trained the unsupervised alignment using adversarial training. \citet{artetxe2018robust} first constructed a low-quality seed dictionary using the assumption that the two vector spaces are isometric and then iteratively updated the mapping and dictionary until convergence. 

\citet{artetxe2018generalizing} comprehensively summarize existing linear methods, showing that the state-of-the-art linear alignment methods can be summarized as an orthogonal mapping. The difference between various methods is solely due to different approaches to vector manipulation (such as normalization, whitening, etc.) before the mapping extraction.

\citet{nakashole2018characterizing} show that, in a small neighbourhood, linear mapping methods work well; however, the linearity assumption does not hold in general, especially for distant languages \cite{ormazabal-etal-2019-analyzing}. A few nonlinear alignment methods have been proposed. \citet{lu2015dcca} trained nonlinear mapping using Deep Canonical Correlation Analysis (DCCA) \cite{andrew2013dcca}, which is an expanded version of a linear Canonical Correlation Analysis (CCA) method, using deep neural networks. They showed that DCCA performs better than linear CCA. Recently, \citet{zhao2020kcca} proposed a nonlinear mapping method, using kernel CCA (KCCA). KCCA projects the vectors into a higher dimensional space and then performs CCA in the new vector space. \citet{zhao2020kcca} report that DCCA has to fine-tune many hyper-parameters and show that KCCA outperforms both DCCA and CCA, especially when data is scarce. In contrast, \citet{lu2015dcca} observe that DCCA scales better with data size than KCCA.

\citet{conneau2018word} used the adversarial training based on generative adversarial networks (GAN) to train a linear mapping between vector spaces. \citet{yang-etal-2018-improving} have used full GAN models for neural machine translation. \citet{fu-etal-2020-absent} trained a bidirectional GAN for cross-lingual alignment of sentence embeddings, improving the results over linear and nonlinear state-of-the-art methods on the sentence alignment task.

In contrast to the above works which deal with non-contextual embeddings, our work is focused on contextual ELMo embeddings. While we use some of the same techniques (GANs for non-linear mappings and Procrustes approach for linear mappings), the task we tackle is different and much more difficult as we have to find a mapping for each context a word appears (and not just each word). The work on contextual mappings, closer to our approach, are outlined below.

\subsection{Related work on contextual mappings}
\label{sec:relatedContextual}
All the above work only concerned static embeddings, not dynamic, contextual embeddings. \citet{schuster2019cross} produced cross-lingual alignments of contextual ELMo embeddings. While each occurrence of a word in contextual embeddings is represented by a different vector, \citet{schuster2019cross} hypothesized that these vectors form clusters. Based on this assumption, they assigned each word a single static vector by calculating the average vector of all word occurrences in a large corpus. They used a linear MUSE method to calculate the alignments of the averaged vectors. This approach's problem is the assumption of isomorphic spaces and loss of information if this assumption is not true in the local context. 

\citet{aldarmaki2019context} used parallel corpora to produce the embedding vectors. They aligned the corpora on the word level, using Fast Align \cite{dyer2013simple}, calculated the ELMo embeddings on the aligned corpora, and extracted a dictionary from the word-level alignments. Their approach showed good results in a sentence translation retrieval task. They measured the accuracy of retrieving the correct translation from the target side of a test parallel corpus using nearest neighbour search and cosine similarity. They applied their approach to three languages (English, German, Spanish). This approach is similar to the linear mappings applied to ELMo, which we describe in \Cref{sec:linearMaps}. The difference is that we use much larger dictionaries and test on many more language pairs. Our non-linear approach, presented in  Sections \ref{sec:architecture} and \ref{sec:training}, has no counterpart in existing works.

\section{Datasets for alignment of contextual embeddings}
\label{sec:alignmentDatasets}
 This section explains the training datasets produced for cross-lingual alignment of contextual ELMo embeddings. These datasets are essential for both linear and non-linear mappings presented in \Cref{sec:elmogan}. Besides the datasets, we also present the language resources used in their creation.
 
Supervised cross-lingual vector alignment approaches assume the existence of a bilingual dictionary, where each word from the dictionary has its own embedding vector. For static, non-contextual embeddings this is straightforward as one can take any of human- or machine-created dictionaries. In contextual embeddings, word vectors depend on the context words appears in. For every context, a word gets a different vector. \citet{schuster2019cross} approached this by averaging all the vectors of a given word, as described in Section~\ref{sec:relatedwork}. This approach loses some information, as words have multiple meanings. For example, the word ``bark'' can refer to the sound a dog makes, a sailing boat, or the outer part of a tree trunk. Furthermore, two meanings may be represented with one word in one language but with two different words in another language. 

We solve these issues by separately aligning each occurrence of a word. We start with a parallel corpus, aligned on a paragraph level to have matching contexts in two languages. 
Let $i$ indicate the index of a context from a parallel corpus $P$. Let $A$ and $B$ represent the first and the second language in a language pair. Then $P_i^A$ is the $i$-th paragraph/context from corpus $P$ in language $A$. Given a bilingual dictionary $D$, let $j$ indicate the index of a word pair in the dictionary so that the dictionary is composed of pairs ($D_j^A, D_j^B), \forall j\in\|D\|$.

We construct our dataset by parsing the parallel corpus. For each word $a \in P_i^A$, we check whether its lemma appears in $D^A$. If it does, given its dictionary index $j$, we check whether $D_j^B$ is a lemma of any word from $P_i^B$. If it is, we add the tuple 
$$(iD_{j}^A, iD_{j}^B, e(D_{j}^A, P_i^A), e(D_{j}^B, P_i^B))$$ 
to our dataset, where $e(D_{j}^A, P_i^A)$ and $e(D_{j}^B, P_i^B))$ are (ELMo) embeddings of the two dictionary words $D_{j}^A$ and $D_{j}^B$, computed in the context $P_i$ for each of the languages, respectively. We considered at most 20 different contexts of each lemma to not overwhelm the dataset with frequent words (such as stop words). For the lemmatization of the corpora, we used the Stanza tool \citep{qi2020stanza} in all analyzed languages. Note that we only used lemmatized corpora for dictionary look-up; for generating the embedding, we used the non-lemmatized corpora.

As we explained in \Cref{sec:elmo}, ELMo models are deep neural networks with three hidden layers. The first layer is non-contextual CNN, followed by two contextual biLSTM layers. The final embedding vectors are constructed from vectors of all three layers.
The first vector is contextually independent, while the second and third layers are contextually dependent.
In the proposed cross-lingual alignment approaches for ELMo, we align vectors from each of the three layers separately. Thus, we need a separate dataset for each layer, i.e., $iD_{j}^A, iD_{j}^B$ components of the learning tuples are the same in the datasets but $e(D_{j}^A, P_i^A), e(D_{j}^B, P_i^B)$ components are computed separately for each contextual LSTM layer.
We created two such contextual datasets for each language pair, one for each of the contextual ELMo layers. For the non-contextual CNN layer, we produce embeddings for every word pair in the bilingual dictionary. As the non-contextual ELMo vectors are the same for all word contexts, the size of this dataset is identical to the bilingual dictionary size.

We split the created datasets into a training and evaluation part. We separately split data for each language pair and each ELMo layer. The training part has 98.5\% of word vector pairs, and the evaluation part has 1.5\% of word vector pairs.

In our work, we considered eleven language pairs from nine different languages. The language pairs along with the sizes of bilingual dictionaries, parallel corpora, and the final training dataset are presented in Table~\ref{tab:contextdict}. For English, we used the original English 5.5B ELMo model\footnote{\url{https://allennlp.org/elmo}}. For Russian, we used the ELMo model trained by DeepPavlov\footnote{\url{https://github.com/deepmipt/DeepPavlov}} on the Russian WMT News. For other seven languages, we used ELMo models trained by \citet{ulcar2020high}\footnote{\url{http://hdl.handle.net/11356/1277}}.

\begin{table}[htb]
\caption{The sizes of dictionaries and parallel corpora used in the creation of ELMo contextual mapping datasets, as well as the size of the resulting datasets.  The sizes of dictionaries are reported in the number of word pairs, the sizes of parallel corpora in the number of matching contexts, and the sizes of resulting datasets in the number of matched words in matched sentence pairs. The Type column describes the dictionary creation approach; ``direct`` and ``triang`` denote that the dictionary was created directly from Wiktionary or with triangulation via English, respectively; ``OES`` stands for the Oxford English-Slovene dictionary. }
\centering
\begin{tabular}{lcrrr}
Language pair & Type & Dictionary & Parallel corpus & ELMo dataset \\ \hline 
English-Estonian & direct & \num{11022} & \num{12486898} & \num{77800} \\
English-Finnish & direct & \num{89307} & \num{27281566} & \num{283000} \\
English-Croatian & direct & \num{3448} & \num{35131729} & \num{44800} \\ 
English-Lithuanian & direct & \num{13960} & \num{1415961} & \num{62800} \\
English-Latvian & direct & \num{10224} & \num{519553} & \num{43800} \\
English-Russian & direct & \num{103850} & \num{25910105} & \num{363800} \\
English-Slovenian & direct & \num{9634} & \num{19641457} & \num{89800} \\
English-Slovenian & OES & \num{182787} & \num{19641457} & \num{294318} \\
English-Swedish & direct & \num{51961} & \num{17660152} & \num{270000} \\
Estonian-Finnish & direct & \num{2191} & \num{9504879} & \num{12800} \\
Estonian-Finnish & triang & \num{43313} & \num{9504879} & \num{78200} \\
Croatian-Slovenian & direct & \num{266} & \num{15636933} & \num{3400} \\
Croatian-Slovenian & triang & \num{3669} & \num{15636933} & \num{31600} \\
Lithuanian-Latvian & direct & \num{2478} & \num{219617} & \num{11200} \\
Lithuanian-Latvian & triang & \num{14545} & \num{219617} & \num{28200} \\
\hline
\end{tabular} 
\label{tab:contextdict}
\end{table}

As sources of parallel texts, we used OpenSubtitles parallel corpora\footnote{\url{https://www.opensubtitles.org/}} \citep{lison2016opensubtitles} from the Opus web page\footnote{\url{http://opus.nlpl.eu}} for each pair of languages. The dictionaries we used are bilingual dictionaries extracted from Wiktionary, using the wikt2dict\footnote{\url{https://github.com/juditacs/wikt2dict}} tool \citep{acs2014wikt}. The tool allows for direct dictionary extraction, as well as triangulation via a third language. In the triangulation case, given three languages, $A$, $B$ and $C$, one constructs a bilingual dictionary for languages $A$ and $B$ so that for every word $a\in A$, one finds its translation $c\in C$ from $A-C$ dictionary. Next, the translation of the word $c$ in language $B$ is found in the $C-B$ dictionary and labeled $b$. The dictionary created using triangulation consists of pairs $a-b$. 

The dictionaries made using the wikt2dict tool are noisy, so we manually filtered them. We replaced the accented vowels with their non-accented variants in languages that do not use accented letters for vowels (e.g., Slovene and Russian). We removed the extra non-alphabetic characters, such as hash symbol, brackets, pipe, etc. We also removed all the entries which contained multiple-word terms. We leave the extension to the alignment of multi-word terms for further work.

We used direct bilingual dictionaries for all language pairs, where one of the languages was English. We used direct dictionaries and dictionaries created with the triangulation via English for all the other pairs (i.e. if neither language is English). For the English-Slovene pair, we also tested a large, high quality, handmade, proprietary Oxford English-Slovene dictionary.

\section{Contextual alignments}
\label{sec:elmogan}
This section describes the proposed methods for cross-lingual alignment of ELMo contextual embeddings. We start with the ELMoGAN method for nonlinear alignment. In Section~\ref{sec:architecture}, we describe the architecture of the proposed alignments, followed by their training in \Cref{sec:training}. Based on the constructed contextual alignment datasets, it is also possible to align contextual embeddings with classical linear mappings. We describe this approach in \Cref{sec:linearMaps}.

\subsection{Architecture of ELMoGAN}
\label{sec:architecture}
The proposed ELMoGAN nonlinear cross-lingual contextual embedding alignment method uses Generative Adversarial Networks (GANs) \citep{goodfellow2014generative}. GANs consist of two connected neural models, a generator and a discriminator. The two models are trained simultaneously via an adversarial process. The discriminator attempts to discern whether the data passed to its input is real or fake (i.e. artificially generated). At the same time, the generator attempts to generate artificial data, which can fool the discriminator. GANs play a zero-sum game, where the discriminator's success means the generator's failure and vice versa. By simultaneously training both networks, they both improve. GANs are mostly used on images, where the described process can lead to compelling new generated images. 

Following the success of GANs in neural machine translation \citep{yang-etal-2018-improving} and cross-lingual embeddings alignment \citep{Conneau2018,fu-etal-2020-absent}, we propose a novel supervised nonlinear mapping method using bidirectional GANs. 
We based our contextual alignment method, called ELMoGAN, on the model of \citet{fu-etal-2020-absent}. Contrary to \citet{fu-etal-2020-absent}, who only used their method with non-contextual fastText embeddings \citep{Bojanowski2017} to align sentences, we align contextual ELMo embeddings \citep{Peters2018}, which is only possible by constructing a special contextual mapping datasets, described in \Cref{sec:alignmentDatasets}. As these datasets encode words in context, they are much larger and the resulting GANs have to more precisely map between cross-lingual vector spaces compared to non-contextual mappings. 

\begin{figure}[htb]
    \centering
    \includegraphics[width=0.7\textwidth]{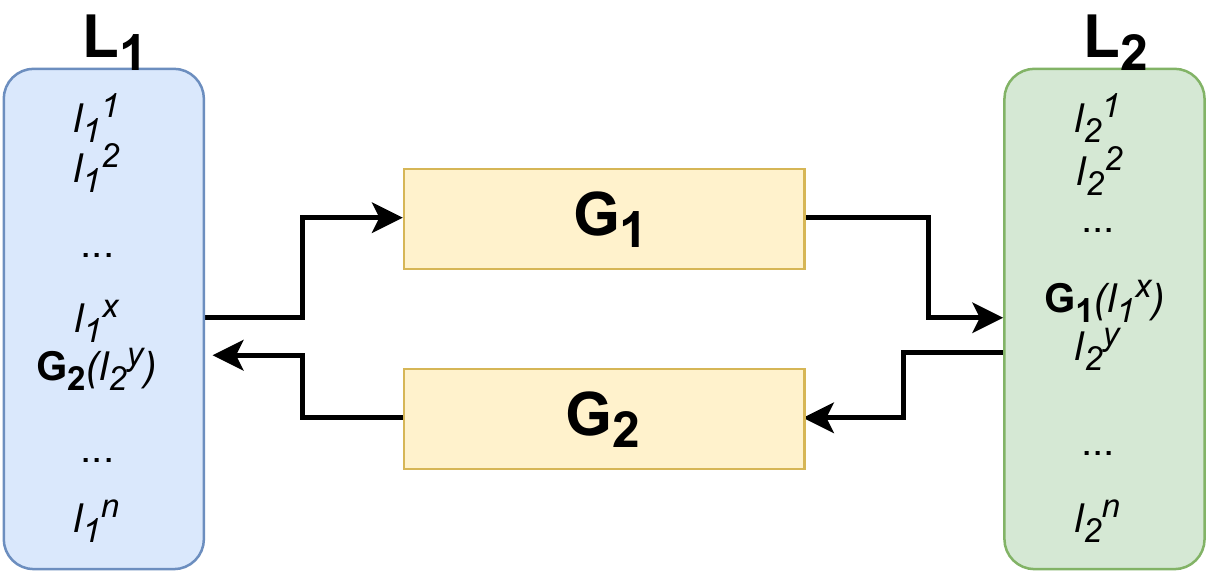}
    \caption{The schema of the two generators for sentence alignment. Generator $G_1$ maps from $L_1$ to $L_2$ and generator $G_2$ vice-versa. For a given pair of matching words $x$ and $y$, where $x$ is from $L_1$ and $y$ from $L_2$, the generator $G_1$ attempts to map the vector $l_1^x$, so that its output $G_1(l_1^x)$ is as close as possible to $l_2^y$. The generator $G_2$ attempts to map $l_2^y$, so that $G_2(l_2^y)$ is as close as possible to $l_1^x$.} 
    \label{fig:elmogan-generator}
\end{figure}

\begin{figure}[h!tb]
    \centering
    \includegraphics[width=0.7\textwidth]{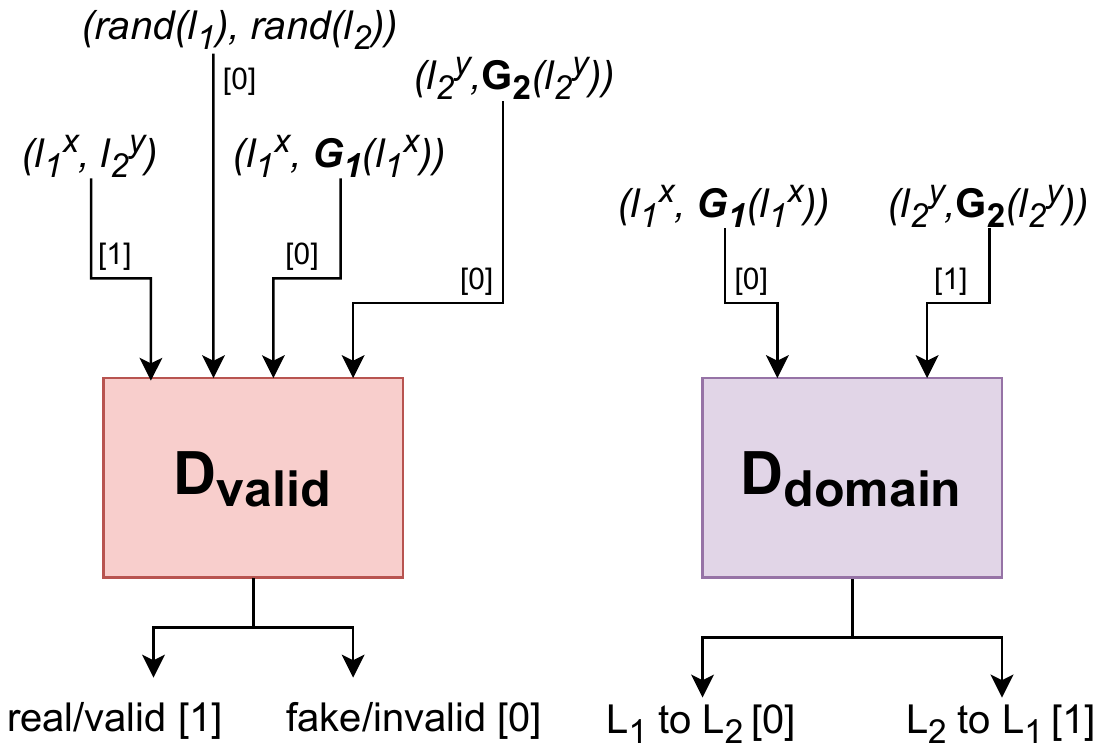}
    \caption{The schema of the two discriminators for sentence alignment. Each discriminator receives a pair of vectors on input. During training $D_{valid}$ receives a pair $(l_1^x, l_2^y)$ labelled as real pair, a pair of random word vectors labelled as fake pair, and outputs from generators labelled as fake pairs. The $D_{domain}$ discriminator receives outputs from generators on input and attempts to discern between them.} 
    \label{fig:elmogan-discriminator}
\end{figure}

The cross-lingual embedding mapping GAN comprises the generator module and discriminator module.
The generator module (\Cref{fig:elmogan-generator}) contains two generators that map vectors from one vector space to the other. Specifically, for a pair of languages $L_1$ and $L_2$, one generator will map from $L_1$ to $L_2$, and the second will map from $L_2$ to $L_1$.  The two generators are completely independent of one another, and they do not share the data during training. The discriminator module (\Cref{fig:elmogan-discriminator}) contains two discriminators. The first discriminator tries to predict whether a given pair of vectors represent the same token, i.e. if the first vector represents the word $x$ in $L_1$ and the second vector represents the translation of the word $x$ in $L_2$ (i.e. $y$). The second discriminator attempts to learn the difference between the direction of mapping. For a given pair of vectors, it predicts whether they are a vector from $L_1$ and its mapping to $L_2$ or a vector from $L_2$ and its mapping to $L_1$.

Compared to the ABSent model by \citet{fu-etal-2020-absent} that was trained on much smaller and easier problem of non-contextual mapping, in ELMoGAN we increased the size of all hidden layers in generators and discriminators. We also significantly lowered the learning rate as we achieved poor results with the learning rate used by \citet{fu-etal-2020-absent}.
The two generators in ELMoGAN have the same architecture: the input layer is followed by three fully connected feed-forward layers of 2048, 4096, and 2048 nodes. We used the ReLU activation function for all three layers. We added a batch normalization layer between each fully connected layer. The output layer has the same size as the input layer. It uses a hyperbolic tangent as the activation function so that the output is between $-1$ and $+1$.
Both discriminators also have the same architecture. We first concatenate the two input vectors, then feed them to the three consecutive fully connected feed-forward layers with leaky ReLU ($\alpha=0.2$). The output layer is a single neuron with sigmoid activation.

\subsection{Training of ELMoGAN}
\label{sec:training}

We jointly trained the generator and discriminator modules using the parallel ELMo vectors datasets, described in Section \ref{sec:alignmentDatasets}. We trained ELMoGAN with the batch size of 256, Adam optimizer with the learning rate of \num{2e-5}, and the learning rate decay of \num{e-5}. For each language pair, we trained three mapping models, one for each of the ELMo layers. For all three models, we used the same settings. 
We feed the generators the word vectors from our training dataset. On the input of the first generator are the vectors from $L_1$, and on the output, there are the matching vectors from $L_2$; vice-versa is true for the second generator.
We feed the first discriminator various pairs of vectors; some represent the same token (True), others represent two different tokens or no token at all (False). For the vector pairs labelled as True, we take the matching pairs from our train dataset. For the vector pairs labelled as False, we have three types of pairs. The first type is two randomly selected vectors from our dataset (one from each vector space). The second type is vectors from $L_1$ and their mappings, using the $G_1$ generator. The third type is vectors from $L_2$ and their mappings, using the $G_2$ generator.

We produced two different versions of ELMoGAN models, based on the number of iterations used in the training. The first version of models (ELMoGAN-10k) were trained for a fixed number of \num{10000} iterations for each layer of each language pair. The second version of models (ELMoGAN-O) were trained models with several different numbers of iterations. We evaluated them on a dictionary induction task and selected the number of iterations that gave the best result. The details of selecting the number of iterations are presented in \Cref{app:elmogano}.

\subsection{Cross-lingual linear mappings for contextual embeddings}
\label{sec:linearMaps}
Besides nonlinear mappings with GANs described above, it is also possible to compute cross-lingual mappings between contextual embeddings based on the standard assumption that the aligned spaces are largely isomorphic. This assumption may be valid for similar languages. Below, we shortly describe methods belonging to this type of alignment approach. 

With a large enough collection of words in matching contexts (as described in \Cref{sec:alignmentDatasets}), we compute their contextual embedding vectors and align them with any of the non-contextual mapping methods.  We use two such well-know and successful methods. The Vecmap methods \citep{artetxe2018generalizing} change both source and target embedding space, which makes them computationally less efficient on downstream tasks as analyzed in \Cref{sec:optimizations}. Methods from the MUSE library \citep{Conneau2018} only align source vectors to target vectors and are therefore computationally more efficient. As discussed in \Cref{sec:relatedContextual}, a similar approach was proposed by \citet{aldarmaki2019context} but did not use large contextual datasets (presented in \Cref{sec:alignmentDatasets}) based on high-quality dictionaries as we did.

\section{Evaluation}
\label{sec:evaluation}
In this section, we compare the four proposed alignment methods for contextual ELMo embeddings, two nonlinear ELMoGAN methods and two linear methods.
We evaluated the methods on four downstream tasks: NER (\Cref{sec:ner}), DP (\Cref{sec:depparsing}), terminology alignment (\Cref{sec:terminology}), and sentiment analysis (\Cref{sec:sentiment}). We compare two nonlinear methods, ELMoGAN-10k and ELMoGAN-O methods, trained as described in Section~\ref{sec:training}, with two linear mapping methods, MUSE \cite{conneau2018word} and Vecmap \cite{artetxe2018robust,artetxe2018generalizing}, adapted for contextual embeddings, as described in \Cref{sec:linearMaps}. These two are linear cross-lingual mapping methods that assume linear transformation between vector spaces. For training of all alignments, we used the same datasets, described in Section~\ref{sec:alignmentDatasets}.
In all the experiments, we use embeddings mappings from a target (i.e evaluation) language to a train language; namely, we map the embeddings of the language used for the evaluation to the vector space of the language, which was used during the training of the model. 

To better interpret the obtained results, we conducted two further ablation studies. In \Cref{sec:datasetSize}, we tested the importance of alignment dataset size. We used the English-Slovene pair, where we have available a large high-quality proprietary Oxford English-Slovene dictionary, instead of the publicly available Wiktionary. In \Cref{sec:optimizations}, we tested different variants of the Vecmap alignment approach to check if we can avoid transforming both the source and target vector space and thereby significantly speed-up the approach.

\subsection{Named Entity Recognition}
\label{sec:ner}
Named entity recognition (NER) is an information extraction task that seeks to locate and classify named entities mentioned in unstructured text into pre-defined categories such as the person names, organizations, locations, medical codes, time expressions, quantities, monetary values, etc. The labels in the used NER datasets are simplified to a common label set of four labels present in all the addressed languages. These labels are "person", "location", "organization", and "other". The latter encompasses all named entities that do not fall in one of the three mentioned classes and all the tokens that are not named entities.
The datasets used in the evaluation on the NER task are shown in Table~\ref{tab:nerdatasets}, along with some basic statistics of the datasets. 

\begin{table}[htb]
	\caption{The collected datasets for the NER task and their properties: number of sentences, number of tagged words, availability, and link to the corpus location.}
	\centering
	\resizebox{\linewidth}{!}{
	\begin{tabular}{llrr}
		Language	& Corpus & Sentences & Tags \\ \hline 
		Croatian	& hr500k \citep{LJUBEI16.340} & 25000 & 29000 \\
		English 	& CoNLL-2003 NER \citep{tjongkimsang2003conll} & 21000 & 44000 \\
		Estonian	& Estonian NER corpus \citep{estonian-ner} & 14000 & 21000 \\
		Finnish  	& FiNER data \citep{ruokolainen2020finer} & 14500 & 17000 \\
		Latvian  	& LV Tagger train data \citep{paikens2012towards} & 10000 & 11500 \\
		Lithuanian  & TildeNER \citep{pinnis-2012-latvian} & 5476 & 7024 \\
		Slovene	& ssj500k \citep{ssj500k} & 9500 & 9500 \\
		Swedish	& Swedish NER \citep{klintberg2015swedishner} & 8500 & 7500 \\
		\hline 
	\end{tabular} }
	\label{tab:nerdatasets}
\end{table}

\begin{table}[h!tb]
\caption{Comparison of different methods for cross-lingual mapping of contextual ELMo embeddings evaluated on the NER task. The best Macro $F_1$ score for each language pair is in bold. The ``Reference`` column represents direct learning on the target language without cross-lingual transfer. The upper part of the table contains a scenario of cross-lingual transfer from English to a less-resourced language, and the lower part of the table shows a transfer between similar languages.} 
\centering
\resizebox{\linewidth}{!}{
\begin{tabular}{lllccccc}
Source. & Target. & Dictionary & Vecmap & ELMoGAN-O & ELMoGAN-10k & MUSE & Reference \\
\hline
English & Croatian & direct & $\mathbf{0.385}$ & 0.274 & 0.365 & 0.024 & 0.810 \\
English & Estonian & direct & 0.554 & 0.693 & $\mathbf{0.728}$ & 0.284 & 0.895 \\
English & Finnish & direct & 0.672 & 0.705 & $\mathbf{0.780}$ & 0.229 & 0.922 \\
English & Latvian & direct & 0.499 & 0.644 & \textbf{0.652} & 0.216 & 0.818 \\
English & Lithuanian & direct & 0.498 & 0.522 & $\mathbf{0.553}$ & 0.208 & 0.755 \\
English & Slovenian & direct & 0.548 & 0.572 & $\mathbf{0.676}$ & 0.060 & 0.850 \\
English & Swedish & direct & \textbf{0.786} & 0.700 & 0.780 & 0.568 & 0.852 \\
\hline
Croatian & Slovenian & direct & 0.387 & 0.279 & 0.250 & \textbf{0.418} & 0.850 \\
Croatian & Slovenian & triang & $\mathbf{0.731}$ & 0.365 & 0.420 & 0.592 & 0.850 \\
Estonian & Finnish & direct & $\mathbf{0.517}$ & 0.339 & 0.316 & 0.278 & 0.922 \\
Estonian & Finnish & triang & $\mathbf{0.779}$ & 0.365 & 0.388 & 0.296 & 0.922 \\
Finnish & Estonian & direct & 0.477 & 0.305 & 0.324 & \textbf{0.506} & 0.895 \\
Finnish & Estonian & triang & \textbf{0.581} & 0.334 & 0.376 & 0.549 & 0.895 \\
Latvian & Lithuanian & direct & $\mathbf{0.423}$ & 0.398 & 0.404 & 0.345 & 0.755 \\
Latvian & Lithuanian & triang & \textbf{0.569} & 0.445 & 0.472 & 0.378 & 0.755 \\
Lithuanian & Latvian & direct & 0.263 & 0.312 & 0.335 & \textbf{0.604} & 0.818 \\
Lithuanian & Latvian & triang & 0.359 & 0.405 & 0.409 & \textbf{0.710} & 0.818 \\
Slovenian & Croatian & direct & 0.361 & 0.270 & 0.307 & \textbf{0.485} & 0.810 \\
Slovenian & Croatian & triang & $\mathbf{0.566}$ & 0.302 & 0.321 & 0.518 & 0.810 \\
\hline
\multicolumn{7}{l}{Average gap for the best cross-lingual transfer in each language} & 0.147 \\
\end{tabular}
}
\label{tab:nerelmo}
\end{table}

In \Cref{tab:nerelmo}, we present the results using the Macro $F_1$ score, which is an average of $F_1$ scores for each class we are trying to predict, excluding the class ``other"  (i.e. not a named entity).
The upper part of \Cref{tab:nerelmo} shows a typical cross-lingual transfer learning scenario, where the model is transferred from resource-rich language (English) to less-resourced languages. In this case, the non-linear ELMoGAN methods, in particular the ELMoGAN-10k variant, are superior to linear Vecmap and MUSE approaches. In this scenario, ELMoGAN-10k is always the best or close to the best mapping approach. This is not the case in the lower part of \Cref{tab:nerelmo}, which shows the second most important cross-lingual transfer scenario: transfer between similar languages. In this scenario, linear Vecmap and MUSE perform best in all twelve language pairs (seven times Vecmap and five times MUSE). We hypothesize that the reason for better performance of linear mappings is the similarity of tested language pairs and therefore lesser violation of the isomorphism assumption the Vecmap and MUSE method make. The results of the MUSE method support this hypothesis. While MUSE performs worst in most cases of transfer from English, the performance gap is smaller for transfer between similar languages. MUSE is sometimes the best method for similar languages, but the results of MUSE fluctuate greatly between language pairs. The second possible factor explaining the results is the quality of the dictionaries, which are in general better for combinations involving English. In particular, dictionaries obtained by triangulation via English are of poor quality, and nonlinear transformations might be more affected by the imprecision of anchor points.

In general, even the best cross-lingual prediction models lag behind the reference model without cross-lingual transfer. The differences in Macro $F_1$ score are small for some languages (e.g., 5.5\% for English-Swedish), but they are significantly larger for most of the languages.

\subsection{Dependency parsing}
\label{sec:depparsing}
The DP task constructs a dependency tree of a given sentence. In DP, all the words in a sentence are arranged into a hierarchical tree, based on their semantic dependencies. Each word has at most one parent node, and only the root word has no parent. A word can have multiple children nodes. In addition to predicting the tree's structure, the task is also to label the hierarchical dependencies.

As the DP architecture, we use the SuPar tool by Yu Zhang\footnote{\url{https://github.com/yzhangcs/parser}}, which is based on the deep biaffine attention \citep{dozat2017deepbiaffine}. We modified the SuPar tool to accept ELMo embeddings on the input; specifically, we used the concatenation of the three ELMo layers. We made the modified code publicly available\footnote{\url{https://github.com/MatejUlcar/parser/tree/elmo}}. We trained the parser for 10 epochs, using datasets in nine languages (Croatian, English, Estonian, Finnish, Latvian, Lithuanian, Russian, Slovene, and Swedish). The datasets are obtained from the Universal Dependencies \citep{universal-dependencies} version 2.3. The datasets used and their basic statistics are shown in Table~\ref{tab:depparse-datasets}.

\begin{table}[htb]
    \centering
	\caption{Dependency parsing datasets and their properties: the treebank, number of sentences, number of tokens, and information about the size of the split.
	}
	\label{tab:depparse-datasets}
	\begin{tabular}{llrrrrr}
	Language & Treebank & Tokens & Sentences & Train & Validation & Test \\
	\hline
	Croatian & SET & $199409$ & $9010$ & $6914$ & $960$ & $1136$ \\
	English & EWT & $254855$ & $16622$ & $12543$ & $2002$ & $2077$ \\
	Estonian & EDT & $438171$ & $30972$ & $24633$ & $3125$ & $3214$ \\
	Finnish & TDT & $202697$ & $15135$ & $12216$ & $1364$ & $1555$ \\
	Latvian & LVTB & $220536$ & $13643$ & $10156$ & $1664$ & $1823$ \\
	Lithuanian & HSE & $5356$ & $263$ & $153$ & $55$ & $55$ \\
	Russian & GSD & $98000$ & $5030$ & $3850$ & $579$ & $601$ \\
	Slovene & SSJ & $140670$ & $8000$ & $6478$ & $734$ & $788$ \\
	Swedish & Talbanken & $96858$ & $6026$ & $4303$ & $504$ & $1219$ \\
	\hline
	\end{tabular}
\end{table}

We used two evaluation metrics in the DP task, the unlabelled and labelled attachment scores (UAS and LAS) on the test set. The UAS and LAS are standard accuracy metrics in DP. The UAS score is defined as the proportion of tokens that are assigned the correct syntactic head. The LAS score is the proportion of tokens assigned the correct syntactic head and the correct dependency label \citep{speech-and-language-processing}.

\begin{table}[htb]
\caption{Comparison of different contextual cross-lingual mapping methods on the DP task. Results are reported as unlabelled attachments score (UAS) and labelled attachment score (LAS). The column ``Direct`` stands for direct learning on the target (i.e. evaluation) language without cross-lingual transfer. The languages are represented with their \href{https://en.wikipedia.org/wiki/List_of_ISO_639-1_codes}{international language codes ISO 639-1}.}
\centering
\resizebox{\linewidth}{!}{
\begin{tabular}{llc|rr|rr|rr|rr|rr}
Train & Eval. &  & \multicolumn{2}{c}{Vecmap} & \multicolumn{2}{c}{ELMoGAN-O} & \multicolumn{2}{c}{ELMoGAN-10k} & \multicolumn{2}{c}{MUSE} & \multicolumn{2}{c}{Direct}\\
lang. & lang. & Dict. & UAS & LAS & UAS & LAS & UAS & LAS & UAS & LAS & UAS & LAS \\ \hline
en & hr & direct & \textbf{73.96} & \textbf{60.53} & 68.73 & 50.29 & 69.74 & 40.93 & 71.01 & 54.89 & 91.74 & 85.84 \\
en & et & direct & \textbf{62.08} & \textbf{40.62} & 52.01 & 30.22 & 44.80 & 24.59 & 58.76 & 34.07 & 89.54 & 85.45 \\
en & fi & direct & \textbf{64.40} & \textbf{45.32} & 50.80 & 25.23 & 42.65 & 22.66 & 55.03 & 37.61 & 90.83 & 86.86 \\
en & lv & direct & \textbf{77.84} & \textbf{65.97} & 68.51 & 49.47 & 67.09 & 39.41 & 76.26 & 63.45 & 88.85 & 82.82 \\
en & lt & direct & \textbf{67.92} & \textbf{39.62} & 58.87 & 28.30 & 57.36 & 21.13 & 66.04 & 37.74 & 55.05 & 24.39 \\
en & ru & direct & \textbf{72.00} & \textbf{16.62} & 60.74 & 8.92 & 60.68 & 8.18 & 65.23 & 14.77 & 89.33 & 83.54 \\
en & sl & direct & \textbf{79.01} & \textbf{59.84} & 68.82 & 48.20 & 67.04 & 43.34 & 77.18 & 56.53 & 93.70 & 91.39 \\
en & sv & direct & 82.08 & 72.74 & 74.39 & 59.70 & 73.81 & 59.63 & \textbf{82.17} & \textbf{72.78} & 89.70 & 85.07 \\ \hline
hr & sl & direct & \textbf{85.47} & \textbf{72.70} & 51.88 & 31.50 & 53.68 & 33.40 & 83.45 & 69.08 & 93.70 & 91.39 \\
hr & sl & triang & \textbf{87.70} & \textbf{76.51} & 54.34 & 36.32 & 59.61 & 38.83 & \textbf{87.70} & 76.40 & 93.70 & 91.39 \\
et & fi & direct & \textbf{79.14} & \textbf{66.09} & 55.67 & 36.85 & 51.35 & 30.66 & 76.66 & 60.01 & 90.83 & 86.86 \\
et & fi & triang & \textbf{80.94} & \textbf{67.35} & 52.63 & 29.94 & 52.83 & 28.70 & 76.96 & 63.37 & 90.83 & 86.86 \\
fi & et & direct & \textbf{75.81} & 57.32 & 54.69 & 33.99 & 53.27 & 32.28 & 74.96 & \textbf{58.14} & 89.54 & 85.45 \\
fi & et & triang & \textbf{79.04} & \textbf{61.86} & 53.64 & 32.73 & 53.86 & 30.13 & 76.74 & 60.27 & 89.54 & 85.45 \\
lv & lt & direct & \textbf{72.38} & \textbf{51.43} & 60.95 & 38.10 & 63.24 & 36.19 & 67.62 & 50.48 & 55.05 & 24.39 \\
lv & lt & triang & \textbf{75.24} & 50.48 & 62.48 & 38.48 & 63.62 & 36.19 & 74.29 & \textbf{53.33} & 55.05 & 24.39 \\
lt & lv & direct & \textbf{63.68} & \textbf{25.88} & 43.50 & 11.54 & 50.70 & 13.69 & 61.05 & 18.87 & 88.85 & 82.82 \\
lt & lv & triang & \textbf{61.86} & \textbf{25.94} & 49.24 & 13.31 & 51.91 & 13.89 & 57.95 & 17.45 & 88.85 & 82.82 \\
sl & hr & direct & \textbf{77.89} & \textbf{62.58} & 47.34 & 29.39 & 52.27 & 32.48 & 72.87 & 55.70 & 91.74 & 85.84 \\
sl & hr & triang & \textbf{81.32} & \textbf{67.51} & 50.96 & 32.82 & 56.17 & 35.96 & 78.63 & 63.96 & 91.74 & 85.84 \\
\hline
\multicolumn{11}{l|}{Average gap for the best cross-lingual transfer in each language} & 6.56 & 17.93 \\
\multicolumn{11}{l|}{Average gap for the best cross-lingual transfer in each language (without Lithuanian)} & 10.38 & 24.62 \\
\end{tabular}
}
\label{tab:depparseresults}
\end{table}

The results in \Cref{tab:depparseresults} show that the Vecmap mapping method outperforms both ELMoGAN methods on all language pairs in this task. Larger dictionaries, created with triangulation, performed better than smaller direct dictionaries, despite the triangulated dictionaries being of worse quality. Language pairs with similar languages performed better than when the training language was English. The exception is the evaluation on Latvian, where the model trained on English performed better than the model trained on Lithuanian. For evaluation on Lithuanian, both models, trained on English and Latvian, outperform the Lithuanian model. This indicates a poorly trained Lithuanian model, which explains the aforementioned exception in the evaluation of Latvian. The Lithuanian model's low performance can be partially explained by the small size of the Lithuanian treebank dataset, as seen in Table~\ref{tab:depparse-datasets}.

The MUSE method is stable on the DP task, which is not the case on the NER task. MUSE performs on par with Vecmap on a few language pairs. Still, its results lie somewhere between Vecmap and ELMoGAN on average.

\subsection{Terminology alignment}
\label{sec:terminology}
Terms are single words or multi-word expressions denoting concepts from specific subject fields. The bilingual terminology alignment task aligns terms between two candidate term lists in two different languages. 
Given a pair of terms $t_1$ and $t_2$, where $t_1$ is from one language and $t_2$ is its equivalent from the second language, we measured the cosine distance between vector of $t_1$ and vectors of all terms from the second language. If the vector of $t_2$ is the closest to $t_1$ among all the terms from the second language, we count the pair as correctly aligned. For example, for a pair of terms from the Slovenian-English term bank ``računovodstvo - accounting'', we map the Slovene word embedding of the word ``računovodstvo'' from Slovene to English and check among all English word vectors for the vector that is the closest to the mapped Slovenian vector for ``računovodstvo''. If the closest vector is ``accounting'', we count this as a success, else as a failure. This measure is called accuracy@1 score or 1NN score and is defined as the number of successes divided by the number of all examples, in this case term pairs. 


We extracted aligned bilingual term lists from Eurovoc \citep{steinberger2002cross}, a multilingual thesaurus with more than 10 thousand terms available in all EU languages.
For building contextualised vector representations of these terms, we used the Europarl corpus \citep{koehn2005europarl,tiedmann2012}. For Croatian, Europarl is not available, so we used DGT translation memory \citep{steinberger2013dgt} instead. 

For each word, we concatenate the three ELMo vectors into one 3072-dimensional vector. For single-word terms, we represent each term as the average vector of all contextual vector representations for that word, found in the corpus. For multi-word terms, we used a two-step approach. First, we check whether the term appears in the corpus. If it does, we represent each term occurrence as the average vector of the words it is composed of. We then average over all the occurrences, as with single-word terms. If the term does not appear in the corpus, we represent it as the average of all words it is composed of, where word vectors are averaged over all occurrences in the corpus. For each language pair, we evaluate the terminology alignment in both directions. That is, given the terms from the first language (source), we search for the equivalent terms in the second language (target), then we repeat the process in the other direction.

We present the results of cross-lingual terminology alignment of contextual ELMo embeddings in \Cref{tab:results-elmo-xlingual-terminology}. We compared the same four mapping methods as in the previous tasks.  For terminology alignment between English and other languages, the two non-linear mappings perform the best on all language pairs. With English as the target language, ELMoGAN-10k always performs the best. In cases where English is the source language, ELMoGAN-O is usually the best.

\begin{table}[h!]
\caption{Comparison of contextual cross-lingual mapping methods for ELMo embeddings, evaluated on the terminology alignment task. Results are reported as accuracy@1, based on the cosine distance metric. The best results for each language and type of transfer (from English or similar language) are typeset in \textbf{bold}. The languages are represented with their \href{https://en.wikipedia.org/wiki/List_of_ISO_639-1_codes}{international language codes ISO 639-1}. The labels ''Dict'' and ''Dir'' in the third column represent the type of the dictionary and the direction of vector mappings: from$\rightarrow$to.}
\centering
\resizebox{0.9\linewidth}{!}{
\begin{tabular}{llc|cccc}
Source & Target & Dict (Dir) & Vecmap & EG-O & EG-10k & MUSE \\ \hline
en & sl & direct (sl$\rightarrow$en) & 0.079 & \textbf{0.152} & 0.151 & 0.096 \\
sl & en & direct (sl$\rightarrow$en) & 0.099 & 0.139 & \textbf{0.195} & 0.126 \\
en & hr & direct (hr$\rightarrow$en) & 0.080 & \textbf{0.153} & 0.135 & 0.116 \\
hr & en & direct (hr$\rightarrow$en) & 0.084 & 0.139 & \textbf{0.153} & 0.102 \\
en & et & direct (et$\rightarrow$en) & 0.092 & \textbf{0.177} & 0.167 & 0.128 \\
et & en & direct (et$\rightarrow$en) & 0.091 & 0.117 & \textbf{0.133} & 0.118 \\
en & fi & direct (fi$\rightarrow$en) & 0.092 & 0.166 & \textbf{0.176} & 0.132 \\
fi & en & direct (fi$\rightarrow$en) & 0.087 & 0.083 & \textbf{0.116} & 0.112 \\
en & lv & direct (lv$\rightarrow$en) & 0.084 & \textbf{0.157} & 0.147 & 0.102 \\
lv & en & direct (lv$\rightarrow$en) & 0.091 & 0.122 & \textbf{0.140} & 0.111 \\
en & lt & direct (lt$\rightarrow$en) & 0.095 & \textbf{0.181} & 0.172 & 0.114 \\
lt & en & direct (lt$\rightarrow$en) & 0.097 & 0.132 & \textbf{0.171} & 0.102 \\
en & sv & direct (sv$\rightarrow$en) & 0.125 & 0.183 & \textbf{0.187} & 0.161 \\
sv & en & direct (sv$\rightarrow$en) & 0.112 & 0.111 & \textbf{0.167} & 0.109 \\
\hline
sl & hr & direct (hr$\rightarrow$sl) & \textbf{0.109} & 0.037 & 0.031 & 0.102 \\
sl & hr & triang (hr$\rightarrow$sl) & 0.130 & 0.056 & 0.046 & \textbf{0.156} \\
sl & hr & direct (sl$\rightarrow$hr) & \textbf{0.109} & 0.039 & 0.038 & 0.100 \\
sl & hr & triang (sl$\rightarrow$hr) & 0.130 & 0.053 & 0.057 & \textbf{0.155} \\
hr & sl & direct (hr$\rightarrow$sl) & \textbf{0.084} & 0.029 & 0.028 & 0.082 \\
hr & sl & triang (hr$\rightarrow$sl) & 0.097 & 0.042 & 0.044 & \textbf{0.121} \\
hr & sl & direct (sl$\rightarrow$hr) & \textbf{0.084} & 0.023 & 0.021 & \textbf{0.084} \\
hr & sl & triang (sl$\rightarrow$hr) & 0.097 & 0.039 & 0.033 & \textbf{0.121} \\
fi & et & direct (et$\rightarrow$fi) & \textbf{0.130} & 0.092 & 0.078 & 0.121 \\
fi & et & triang (et$\rightarrow$fi) & \textbf{0.130} & 0.102 & 0.080 & 0.124 \\
fi & et & direct (fi$\rightarrow$et) & \textbf{0.129} & 0.085 & 0.089 & 0.122 \\
fi & et & triang (fi$\rightarrow$et) & 0.130 & 0.090 & 0.094 & \textbf{0.145} \\
et & fi & direct (et$\rightarrow$fi) & 0.143 & 0.091 & 0.094 & \textbf{0.167} \\
et & fi & triang (et$\rightarrow$fi) & 0.148 & 0.095 & 0.103 & \textbf{0.166} \\
et & fi & direct (fi$\rightarrow$et) & 0.143 & 0.108 & 0.092 & \textbf{0.166} \\
et & fi & triang (fi$\rightarrow$et) & 0.148 & 0.118 & 0.097 & \textbf{0.189} \\
lv & lt & direct (lt$\rightarrow$lv) & 0.102 & 0.080 & 0.061 & \textbf{0.123} \\
lv & lt & triang (lt$\rightarrow$lv) & 0.119 & 0.090 & 0.076 & \textbf{0.134} \\
lv & lt & direct (lv$\rightarrow$lt) & 0.102 & 0.059 & 0.071 & \textbf{0.123} \\
lv & lt & triang (lv$\rightarrow$lt) & 0.119 & 0.065 & 0.077 & \textbf{0.128} \\
lt & lv & direct (lt$\rightarrow$lv) & 0.099 & 0.061 & 0.069 & \textbf{0.102} \\
lt & lv & triang (lt$\rightarrow$lv) & 0.112 & 0.064 & 0.076 & \textbf{0.116} \\
lt & lv & direct (lv$\rightarrow$lt) & 0.099 & 0.071 & 0.057 & \textbf{0.102} \\
lt & lv & triang (lv$\rightarrow$lt) & \textbf{0.112} & 0.083 & 0.069 & 0.110 \\
\hline
\end{tabular}
}
\label{tab:results-elmo-xlingual-terminology}
\end{table}

For terminology alignment between similar languages we also compared the mapping of embeddings in both directions, that is we mapped source terms embeddings to target terms space and vice-versa. Linear methods outperform the non-linear methods on similar languages. In most cases, MUSE is the best method. If we just look at the best dictionary and mapping direction for each language pair, MUSE is the best in each language pair not involving English.
The terminology alignment is in most cases better from English than from a similar language as a source, the exceptions are Croatian and Finnish (as targets).

Overall, it seems that results on the terminology alignment task follow the same pattern as in the NER task: dominance of nonlinear methods in the transfer from English and prevalence of linear methods in the transfer from similar languages.

\subsection{Sentiment analysis}
\label{sec:sentiment}
Our evaluation treats sentiment analysis as a sentence-level classification task. Given a sentence or a short text of a few sentences, we wish to label it with one of the three predefined categories: positive, negative, or neutral, based on the text's sentiment. We used large Twitter sentiment datasets by \citet{mozetivc2016multilingual}, using only four languages we are working with, i.e. English, Croatian, Slovenian, and Russian.

\begin{table}[h!tb]
\caption{Comparison of different methods for cross-lingual mapping of contextual ELMo embeddings, evaluated on the sentiment analysis task. The best macro $F_1$ score for each language pair is in bold. The ``Ref.`` column represents direct learning on the target language without cross-lingual transfer. The upper part of the table contains a scenario of cross-lingual transfer from English to a less-resourced language, and the lower part of the table shows a transfer between similar languages.} 
\centering
\resizebox{\linewidth}{!}{
\begin{tabular}{lllccccc}
Source & Target & Dict. & Vecmap & ELMoGAN-O & ELMoGAN-10k & MUSE & Ref.\\
\hline
English & Croatian & direct & \textbf{0.44} & 0.43 & 0.42 & 0.41 & 0.59 \\
English & Slovenian & direct & \textbf{0.45} & 0.41 & 0.43 & 0.43 & 0.51 \\
English & Russian & direct & \textbf{0.45} & 0.33 & 0.36 & 0.44 & 0.68 \\
\hline
Croatian & Slovenian & direct & \textbf{0.45} & 0.36 & 0.39 & 0.44 & 0.51 \\
Croatian & Slovenian & triang & \textbf{0.46} & 0.37 & 0.40 & 0.43 & 0.51 \\
Slovenian & Croatian & direct & 0.33 & 0.32 & \textbf{0.37} & 0.34 & 0.59 \\
Slovenian & Croatian & triang & \textbf{0.47} & 0.31 & 0.34 & 0.31 & 0.59 \\

\hline
\end{tabular}
}
\label{tab:sentimentelmo}
\end{table}

We trained the classifiers using the same approach and hyper-parameters as for NER (described in \Cref{sec:ner}), but with a different neural network architecture. We trained models with four hidden layers, three bidirectional LSTMs and one fully connected feed-forward layer. The three LSTM layers have 512, 512, and 256 units, respectively. The fully connected layer has 64 neurons.

We present the results using the macro $F_1$ score in \Cref{tab:sentimentelmo}. The difference between transfer from English and transfer from a similar language is much smaller in this task. Similarly, the performance between direct dictionaries and those created using triangulation is not large in most cases. Triangulated dictionaries perform better overall, but not for every mapping method. Vecmap method outperforms the other three methods in all, but one example, transfer from Slovenian to Croatian using direct dictionaries, where ELMoGAN-10k performs the best. When using the triangulated dictionary on the same pair, Vecmap outperforms other methods by a larger margin.

Overall, the results on the sentiment analysis task follow the same pattern as on the DP task: dominance of the linear mapping methods.

\subsection{Dataset size importance}
\label{sec:datasetSize}
We tested the importance of dataset size on the English-Slovene language pair. In the contextual dataset creation, we used a large, high-quality Oxford English-Slovene dictionary instead of Wiktionary. We kept all the other resources and settings the same. We evaluated ELMoGAN-10k on NER and DP tasks using various sizes of the dataset to train contextual alignments. One of the dataset sizes is \num{89800} entries, which is the same size as the dataset created with a low-quality Wiktionary dictionary. We included that size for easier comparison between both dictionaries.
\begin{figure}[p]
\captionlistentry[table]{A table beside a figure}
\captionsetup{labelformat=andtable}\label{tab:nerelmodict}
\caption{Comparison of different sizes of cross-lingual contextual datasets based on different dictionaries used for cross-lingual mapping of contextual ELMo embeddings, evaluated on the NER task. LQsize represents the size of the dataset equal to the size of the low-quality dictionary (\num{89800} total entries, \num{88453} entries in the training dataset). The mapping method used was ELMoGAN-10k.}
\label{fig:nerelmodict}
\centering
\includegraphics[width=0.59\textwidth]{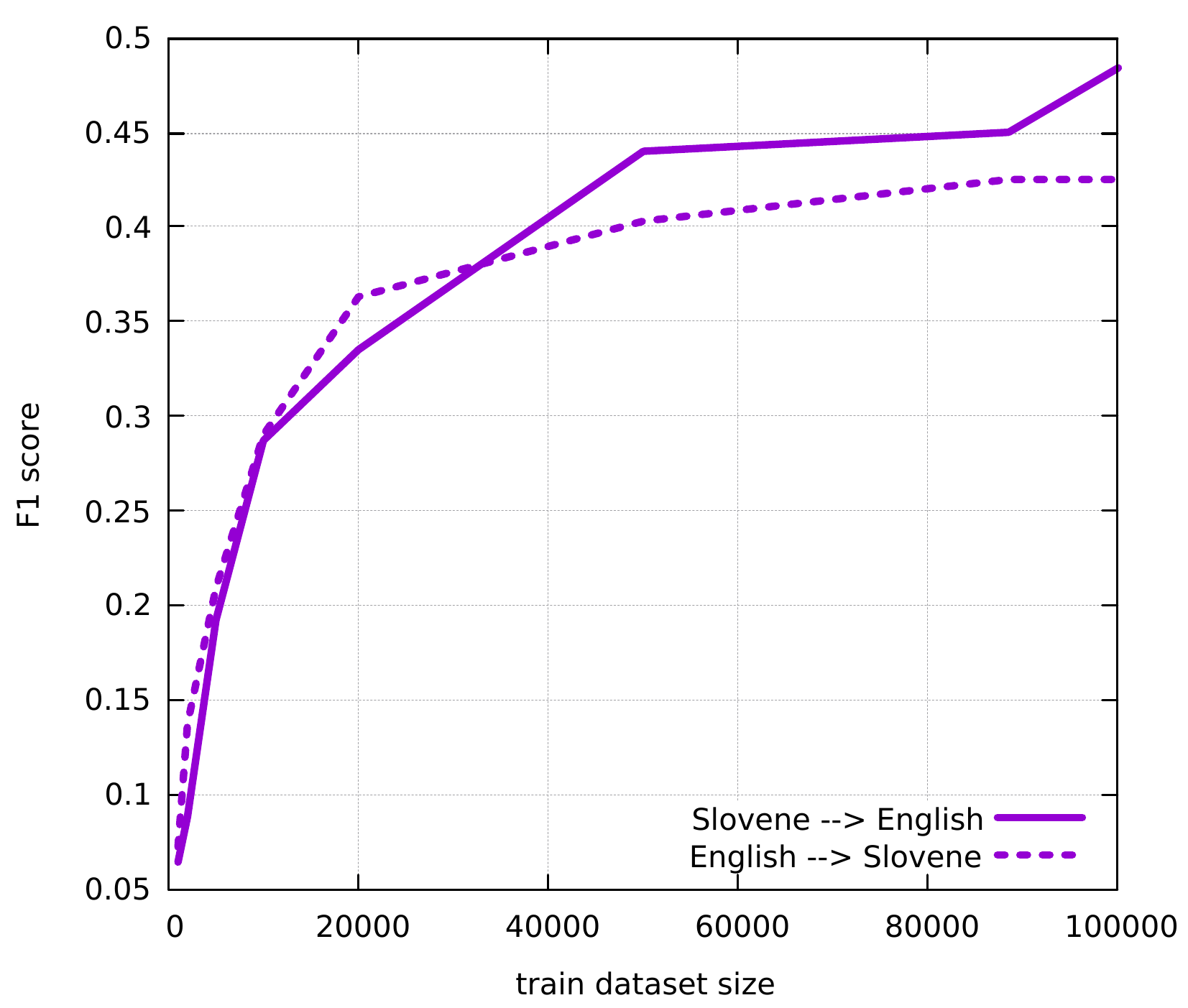}
{
\resizebox{0.39\textwidth}{!}{
\begin{tabular}[b]{llrr}
Train & Eval. & size & macro $F_1$ \\
\hline
en & sl & 1k & 0.064 \\
en & sl & 2k & 0.088 \\
en & sl & 5k & 0.192 \\
en & sl & 10k & 0.287 \\
en & sl & 20k & 0.335 \\
en & sl & 50k & 0.440 \\
en & sl & 100k & \textbf{0.484} \\
en & sl & LQsize & 0.450 \\
\hline
sl & en & 1k & 0.072 \\
sl & en & 2k & 0.137 \\
sl & en & 5k & 0.209 \\
sl & en & 10k & 0.290 \\
sl & en & 20k & 0.363 \\
sl & en & 50k & 0.403 \\
sl & en & 100k & \textbf{0.425} \\
sl & en & LQsize & \textbf{0.425} \\
\hline
\end{tabular}
}}
\end{figure}

The results on the NER task are shown in \Cref{fig:nerelmodict} and \Cref{tab:nerelmodict}. When we increase the size of the dataset, the performance on the NER task improves. The dataset size matters, and we presume that the performance would further increase with an even larger dataset. Surprisingly, the results achieved with the dataset of size \num{89800} are significantly worse than the results achieved with the dataset of the same size, created with a low-quality dictionary (see \Cref{tab:nerelmo}). Using the Oxford English-Slovene dictionary, we achieved the $F_1$ score of $0.450$ when trained on English and evaluated on Slovene. Using Wiktionary bilingual dictionary, we achieved the $F_1$ score of $0.676$ on the same language pair with the same alignment method.

\begin{figure}[p]
\captionlistentry[table]{A table beside a figure}
\captionsetup{labelformat=andtable}\label{tab:dpelmodict}
\caption{
Comparison of different sizes of cross-lingual contextual datasets based on different dictionaries used for cross-lingual mapping of contextual ELMo embeddings, evaluated on the DP task. LQsize represents the size of the dataset based on the low-quality dictionary (\num{89800} total entries, \num{88453} entries in train dataset). We used the ELMoGAN-10k mapping method.}
\label{fig:dpelmodict}
\centering
\includegraphics[width=0.59\textwidth]{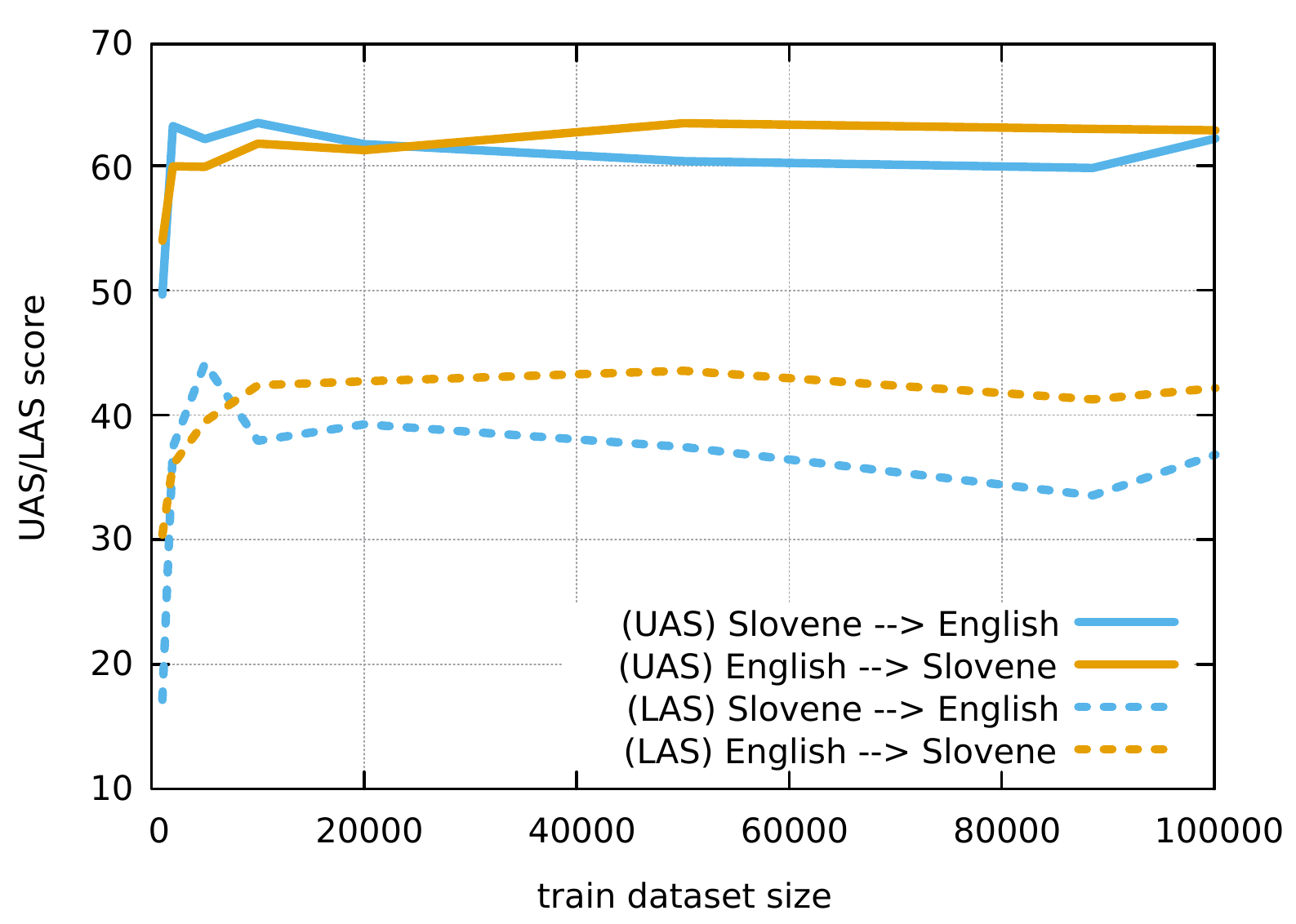}
\resizebox{0.39\textwidth}{!}{
\begin{tabular}[b]{llrrr}
Train & Eval. & size & UAS & LAS \\ \hline
en & sl & 1k & 49.71 & 17.05 \\
en & sl & 2k & 63.28 & 37.40 \\
en & sl & 5k & 62.25 & \textbf{44.16} \\
en & sl & 10k & \textbf{63.53} & 37.94 \\
en & sl & 20k & 61.83 & 39.26 \\
en & sl & 50k & 60.47 & 37.43 \\
en & sl & 100k & 62.28 & 36.81 \\
en & sl & LQsize & 59.90 & 33.53 \\ \hline
sl & en & 1k & 54.06 & 30.35 \\
sl & en & 2k & 60.05 & 36.00 \\
sl & en & 5k & 60.00 & 39.49 \\
sl & en & 10k & 61.89 & 42.41 \\
sl & en & 20k & 61.34 & 42.73 \\
sl & en & 50k & \textbf{63.53} & \textbf{43.58} \\
sl & en & 100k & 62.94 & 42.18 \\
sl & en & LQsize & 63.07 & 41.27 \\ \hline
\end{tabular}
}
\end{figure}

The results in the DP task show different behaviour. At first, the performance quickly increases with larger datasets, and then it stays the same or even starts to drop (see \Cref{fig:dpelmodict} and \Cref{tab:dpelmodict}). The best results are achieved with the dataset of size \num{50000} when mapping from English to Slovene. When mapping from Slovene to English, datasets of size only \num{10000} (based on UAS) and \num{5000} (based on LAS) achieve the best results.

The better performance of Wiktionary bilingual dictionary over high-quality Oxford dictionary, when datasets are of the same size, is observed in the DP task as well. On Slovene to English mapping, the dataset from Oxford dictionary scores $59.90\%$ UAS and $33.53\%$ LAS. Wiktionary-based dataset scores $67.04\%$ UAS and $43.34\%$ LAS. 

The results on dataset size and results from Sections~\ref{sec:ner} and \ref{sec:depparsing} lead us to the conclusion that the quality of the dictionary used does not play a large role. The more important parameter is the size of the dictionary. On the NER task, larger dictionary sizes always improve results. The DP task results remain inconclusive, as the larger dictionary created with triangulation outperformed the smaller direct dictionary for similar language pairs on the DP task.

\subsection{Vecmap optimizations}
\label{sec:optimizations}
In computing cross-lingual alignments of two languages, the Vecmap method changes both embedding spaces. This means that we have to train a separate embedding for each language pair. In our case, we had to train eight different English models on English data for each downstream task, one for each pair of languages, when using Vecmap for alignments. The reason is that the English vectors change during the alignment as well, and we have to apply that change at the time of classifier model training. This considerably slows down the training and evaluating procedure. We tested several approaches to avoid retraining separate models, but none was successful. A detailed description of our experiments is contained in \Cref{app:vecmap}.

\section{Conclusion}
\label{sec:conclusions}
We present four novel methods for cross-lingual mapping of contextual ELMo embeddings. The two ELMoGAN methods (ELMoGAN-O and ELMoGAN-10k) do not assume isomorphic embedding spaces and use GANs to compute the alignments. The two linear methods use Vecmap and MUSE libraries, but map words in matching contexts. To construct the contextual mappings, all four methods need contextual embeddings datasets. We constructed fifteen such datasets for eleven language pairs. 
We created a matching set of contextual word embeddings for each language pair and each ELMo layer from parallel corpora and bilingual dictionaries. 

Nonlinear ELMoGAN methods outperformed linear mappings on the NER and terminology alignment, especially on distant languages where the assumption of isomorphic embedding spaces is strongly violated. The nonlinear methods performed worse on the DP and sentiment analysis tasks. There are substantial differences between languages and we advise users to test both linear and nonlinear methods on a specific task and language.
 The ELMoGAN approach is sensitive to the values of training parameters,  mostly the learning rate and the number of iterations.  To find a set of well-performing hyperparameters, this method has to be carefully fine-tuned for each task. 
 
 There are still some open questions in practical application of ELMoGAN approach, e.g., on the methodology for choosing the right number of iterations for each task. Further, the dictionary induction task we currently use internally to determine the right number of iterations works well for the NER and terminology alignment tasks, but seems inappropriate for the DP task where greater emphasis is on syntactic properties of the language (and not so much on the words as in the NER task).

 In further work, we intend to work on a robust method to find hyper-parameters. Testing more GAN architectures to find a more robust mapping might be a promising research direction. An issue worth investigating is multiple-word terms which are not included in the current contextual mapping datasets but could be very useful in tasks requiring their joint recognition. 

\section*{Acknowledgments} 
The work was partially supported by the Slovenian Research Agency (ARRS) core research programme P6-0411 and project J1-2480.
This paper is supported by European Union's Horizon 2020 research and  innovation programme under grant agreement No 825153, project EMBEDDIA (Cross-Lingual Embeddings for Less-Represented Languages in European News Media).
The results of this publication reflect only the authors' view and the EU Commission is not responsible for any use that may be made of the information it contains.

\bibliographystyle{model5-names}
\bibliography{bibliography}

\appendix

\section{Tuning the number of iterations of ELMoGAN-O}
\label{app:elmogano}
ELMoGAN mapping models have been trained for a different number of iterations for each language pair and each ELMo layer. We have trained all models for the numbers of iterations set between \num{6000} and \num{50000}. We evaluated each model on the dictionary induction task on the evaluation part of our contextual mapping dataset, presented in \Cref{sec:alignmentDatasets}. We have used the average score of precision@1, precision@5, and precision@10 for both directions of our bidirectional mapping model (i.e. from the first to the second language and reverse). The number of iterations that produced the best result on the evaluation set was selected as the optimal and was used in the model called ELMoGAN-O in other evaluations. The selected numbers of iterations are presented in \Cref{tab:elmoganO}. 

\begin{table}[htb]
\caption{The number of iterations ELMoGAN-O was trained for, for each embedding layer and language pair. The optimal number of iterations was determined on the dictionary induction task.} 
\centering
\begin{tabular}{lllrrr}
Language 1 & Language 2 & Dictionary & Layer 1 & Layer 2 & Layer 3 \\
\hline
English & Croatian & direct & 15000 & 50000 & 30000 \\
English & Estonian & direct & 12000 & 50000 & 40000 \\
English & Finnish & direct & 40000 & 50000 & 50000 \\
English & Latvian & direct & 10000 & 40000 & 50000 \\
English & Lithuanian & direct & 30000 & 50000 & 40000 \\
English & Slovenian & direct & 30000 & 50000 & 25000 \\
English & Swedish & direct & 50000 & 50000 & 50000 \\
Croatian & Slovenian & direct & 15000 & 40000 & 10000 \\
Croatian & Slovenian & triangular & 25000 & 50000 & 25000 \\
Estonian & Finnish & direct & 30000 & 25000 & 15000 \\
Estonian & Finnish & triangular & 30000 & 50000 & 20000 \\
Latvian & Lithuanian & direct & 25000 & 40000 & 30000 \\
Latvian & Lithuanian & triangular & 50000 & 30000 & 25000 \\
\hline
\end{tabular}
\label{tab:elmoganO}
\end{table}

\begin{figure}[h!t]
    \centering
    \includegraphics[width=0.9\textwidth]{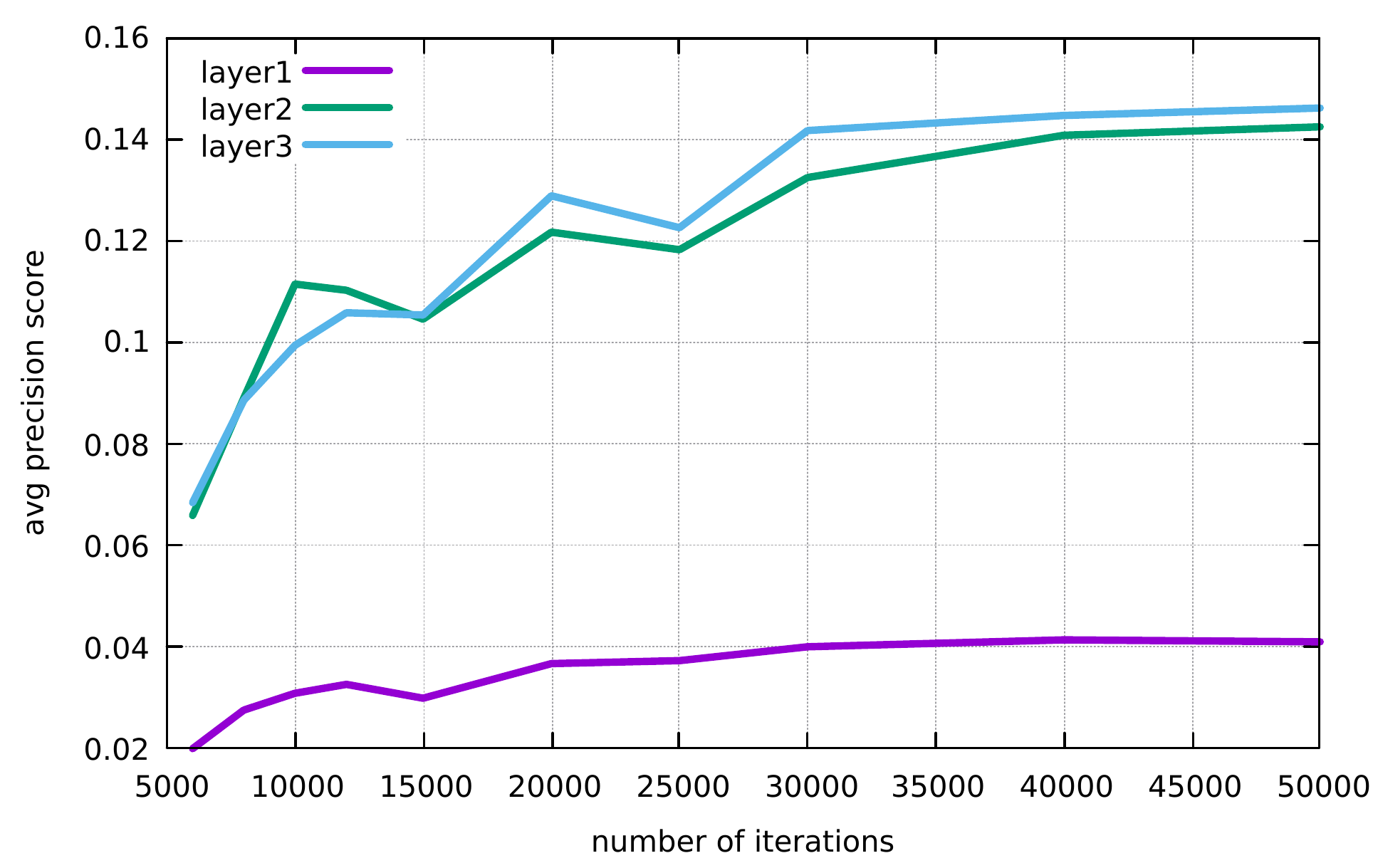}
    \caption{The average precision score on dictionary induction task for English-Finnish alignment at different numbers of iterations of alignment algorithm.} 
    \label{fig:elmoganO}
\end{figure}

We opted not to check more than \num{50000} iterations because the precision on the evaluation task rises quite quickly and then saturates or drops. For example, on the English-Finnish pair, the selected iterations were \num{40000} for layer 1 and \num{50000} for layers 2 and 3. Still, these numbers do not fully reflect the optimal behaviour for all languages and dictionaries. The precision scores for different iterations for this language pair are shown in \Cref{fig:elmoganO}.

\section{Vecmap speed-up experiments}
\label{app:vecmap}
As explained in \Cref{sec:optimizations}, the Vecmap method changes both languages' embedding spaces when computing the cross-lingual alignments. This means that we have to train a separate embedding for each language pair; in our case, we had to train eight different English models, one for each pair of languages. This considerably slows down the training and evaluating procedure. We tested six different sets of options for the Vecmap method to avoid retraining separate models. In this experiment, the training language was always English and we used the DP task. By default, Vecmap first normalizes both vector sets of a language pair and then calculates the mapping matrix, which maps vectors from one language to the other language (in our experiment, from each language to English). Finally, it re-weighs both sets of vectors. In the results below, we denote this approach as ``ELMoVM'' and it is identical to how we used the Vecmap method elsewhere in this paper. We tested five alternative approaches on the DP task; all of them were unsuccessful. This extra step of mapping both source and target languages seems to be unavoidable. The results for all the approaches are shown in Table~\ref{tab:elmovm-alt-results}.

\begin{table}[h!t]
\caption{Various options used with the Vecmap method on the DP task. Train language is always English.}
\centering
\resizebox{\linewidth}{!}{
\begin{tabular}{l|rr|rr|rr|rr|rr|rr}
Eval. & \multicolumn{2}{c}{ELMoVM} & \multicolumn{2}{c}{et} & \multicolumn{2}{c}{orth} & \multicolumn{2}{c}{nonorm} & \multicolumn{2}{c}{evalnorm} & \multicolumn{2}{c}{def} \\
lang. & UAS & LAS & UAS & LAS & UAS & LAS & UAS & LAS & UAS & LAS & UAS & LAS \\ \hline
hr & \textbf{73.96} & \textbf{60.53} & 26.79 & 1.83 & 13.22 & 1.14 & 17.03 & 2.29 & 25.67 & 6.10 & 16.54 & 0.72 \\
et & \textbf{62.08} & \textbf{40.62} & 62.08 & 40.62 & 11.97 & 1.21 & 9.38 & 0.76 & 20.05 & 1.62 & 13.82 & 1.53 \\
fi & \textbf{64.40} & \textbf{45.32} & 11.38 & 0.76 & 15.41 & 0.49 & 18.11 & 0.62 & 24.50 & 1.94 & 18.53 & 0.83 \\
lv & \textbf{77.84} & \textbf{65.97} & 25.32 & 2.21 & 12.82 & 1.26 & 12.88 & 0.63 & 29.10 & 7.89 & 20.39 & 1.96 \\
lt & \textbf{67.92} & \textbf{39.62} & 9.43 & 0.00 & 7.55 & 0.00 & 7.55 & 0.00 & 15.09 & 0.00 & 11.32 & 1.89 \\
sl & \textbf{79.01} & \textbf{59.84} & 28.92 & 2.53 & 13.72 & 0.87 & 12.06 & 0.48 & 25.22 & 6.45 & 14.33 & 0.83 \\
sv & \textbf{82.08} & \textbf{72.74} & 26.23 & 5.23 & 13.50 & 1.46 & 11.27 & 0.81 & 26.45 & 11.92 & 15.22 & 1.41 \\ \hline
\end{tabular}
}
\label{tab:elmovm-alt-results}
\end{table}

\begin{table}[htb]
\caption{Options used (y) or not used (n) for different variants of Vecmap mapping.}
\centering
\resizebox{\linewidth}{!}{
\begin{tabular}{l|cccccc}
Method & ELMoVM & et & orth & nonorm & evalnorm & def \\
\hline
Train lang. mapped & y & y & n & n & n & n \\
Normalization at train time & y & y & n & n & n & n \\
Eval lang. mapped & y & y & y & y & y & y \\
Normalization at eval time & y & y & n & n & y & y \\
Normalization used for mapping calc. & y & y & y & n & n & y \\ \hline
\end{tabular}
}
\label{tab:elmovm-alt-options}
\end{table}

 The options for all the approaches are summarized in Table~\ref{tab:elmovm-alt-options}. The approach ``et'' is identical to ELMoVM, except that we used the English model trained for Estonian-English pair for all language pairs. The following four approaches do not alter English vectors in any way. Approach ``orth'' removes the normalization performed during the evaluation, but the normalization was still used for both languages to calculate the mapping matrix. Method ``nonorm'' is identical to ``orth'', except that we removed the normalization also during the mapping matrix calculation. Method ``evalnorm'' adds the normalization during the evaluation but does not use it during the mapping matrix calculation. Finally, the approach ``def'' uses the normalization both during the evaluation and mapping matrix calculation.

\end{document}